\documentclass[10pt]{article}
\usepackage[utf8]{inputenc}
\usepackage[T1]{fontenc}
\usepackage[a4paper,margin=0.8in]{geometry}
\usepackage{xcolor}
\usepackage{multirow}
\usepackage{booktabs}
\usepackage{graphicx}
\usepackage[square,numbers]{natbib}
\usepackage{authblk}
\usepackage{url}
\usepackage{hyperref}
\usepackage{amsmath}

\usepackage{fancyhdr}
\definecolor{wind}{HTML}{337395}
\definecolor{solar}{HTML}{f8a000}

\title{Global Renewables Watch: A Temporal Dataset of Solar and Wind Energy Derived from Satellite Imagery}

\author[1]{Caleb Robinson\thanks{These authors contributed equally to this work. \\ Corresponding authors: \texttt{caleb.robinson@microsoft.com}, \texttt{anthony.ortiz@microsoft.com}}}
\author[1]{Anthony Ortiz$^*$}
\author[1]{Allen Kim}
\author[1]{Rahul Dodhia}
\author[3]{Andrew Zolli}
\author[2]{Shivaprakash K Nagaraju}
\author[2]{James Oakleaf}
\author[2]{Joe Kiesecker}
\author[1]{Juan M. Lavista Ferres}

\affil[1]{Microsoft AI for Good Research Lab, Redmond, WA, USA}
\affil[2]{The Nature Conservancy (TNC), Washington D.C., USA}
\affil[3]{Planet Labs PBC, San Francisco, CA, USA}

\date{}

\begin{document}

\maketitle

\begin{abstract}
We present a comprehensive global temporal dataset of commercial solar photovoltaic (PV) farms and onshore wind turbines, derived from high-resolution satellite imagery analyzed quarterly from the fourth quarter of 2017 to the second quarter of 2024. We create this dataset by training deep learning-based segmentation models to identify these renewable energy installations from satellite imagery, then deploy them on over 13 trillion pixels covering the world. For each detected feature, we estimate the construction date and the preceding land use type. This dataset offers crucial insights into progress toward sustainable development goals and serves as a valuable resource for policymakers, researchers, and stakeholders aiming to assess and promote effective strategies for renewable energy deployment. Our final spatial dataset includes 375,197 individual wind turbines and 86,410 solar PV installations. We aggregate our predictions to the country level — estimating total power capacity based on construction date, solar PV area, and number of windmills — and find an $r^2$ value of $0.96$ and $0.93$ for solar PV and onshore wind respectively compared to IRENA's most recent 2023 country-level capacity estimates.
\end{abstract}

\begin{figure*}[hb]
    \centering
    \includegraphics[width=1\textwidth]{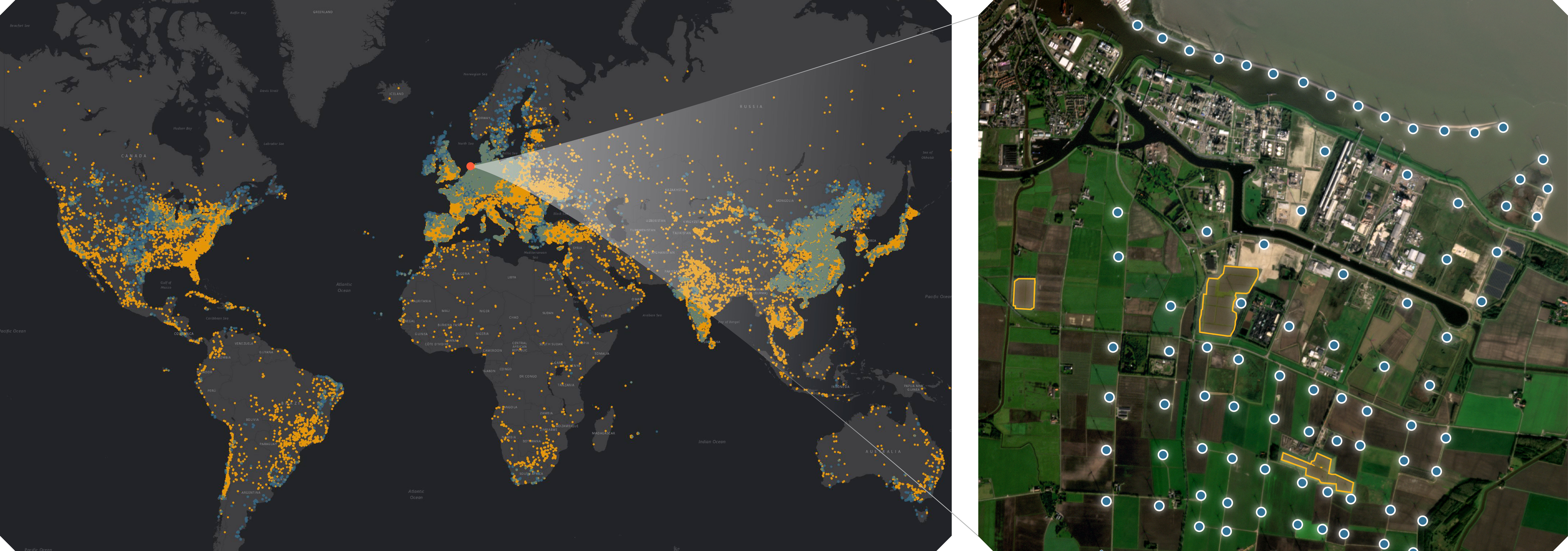}
    \caption{(\textbf{Left}) Distribution of 86,410 \textcolor{solar}{solar PV installations} and 375,197 \textcolor{wind}{onshore windmills} detected by our models in 2024 Q2. (\textbf{Right}) Close-up visualizations of solar and wind installations in the village of Farmsum in the Dutch province of Groningen.}
    \label{fig:1}
\end{figure*}

Climate change poses significant challenges globally, manifesting through shifting weather patterns, threats to food production, and rising sea levels that increase the risk of catastrophic flooding~\cite{ipcc2023synthesis}. The energy sector, responsible for approximately three-quarters of global greenhouse gas emissions, plays a pivotal role in these environmental changes~\cite{anika2022prospects}. To mitigate the most severe impacts of climate change, a transformation toward renewable energy sources is essential. Notably, 87\% of Nationally Determined Contributions (NDCs) under the Paris Climate Accord aim to increase renewable energy usage, with over half specifying quantifiable targets\footnote{\url{https://www.climatewatchdata.org/net-zero-tracker}}.

Monitoring the expansion of renewable energy, particularly wind and solar installations --- which constitute over 80\% of newly installed renewable capacity --- is crucial for meeting climate-related commitments~\cite{wiatrosMotyka2022global}. Further, the rapid development of renewable energy infrastructure necessitates careful consideration of land use, as land is a limited resource in many regions~\cite{chinese2011power}. The physical footprint and siting of these installations can impact ecosystems, cultural and historical resources, scenic landscapes, and agricultural production~\cite{jones2015energy}. 
Proactive spatial planning can guide renewable energy development toward low-conflict areas, balancing economic viability with conservation priorities. Identifying ``Renewables Acceleration Areas'' through robust spatial planning --- including ecosystem and wildlife sensitivity mapping, ecological priorities, and social value assessments --- can facilitate sustainable development~\cite{eu2023}. Analyzing past patterns of renewable energy expansion can also inform predictive models that optimize future siting, minimizing environmental conflicts~\cite{evans2014shale,oakleaf2019mapping,kiesecker2023road}. Therefore, tracking the spatial patterns of renewable energy development is vital to ensure that such projects do not compromise biodiversity and other essential ecosystem services~\cite{kiesecker2017energy,kiesecker2024land}. However, existing data about the location and history of renewable energy installations is limited and existing  datasets become quickly outdated with the exponential growth of solar and wind power. Renewable Energy Capacity grew by more than 80\% in the last 5 years with solar PV and wind accounting for 95\% of the expansion \cite{2024iea}.

\begin{figure}[tb]
    \centering
    \includegraphics[width=1.0\linewidth]{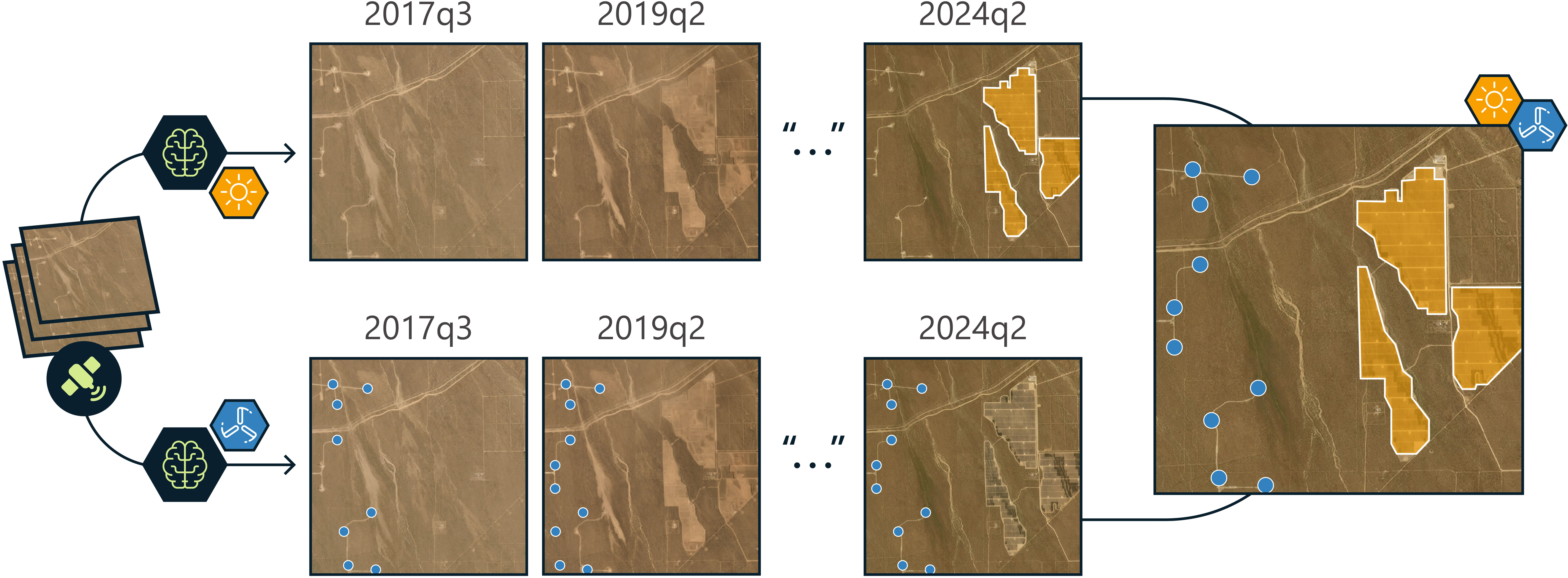}
    \caption{\textbf{Inference workflow}. We run separate solar PV and windmill semantic segmentation models over PlanetScope quarterly satellite imagery from 2017 Q4 through 2024 Q2 to create a solar PV polygon dataset and windmill point dataset with estimated construction dates from each feature.}
    \label{fig:2}
\end{figure}

Global satellite imagery layers combined with machine learning methods for extracting renewable energy locations offer the promise of creating continuously updated datasets of renewable energy installations along with their history and other metadata identifiable from space. Deep learning, in general, is a powerful tool for extracting information from satellite imagery, and Convolutional Neural Networks (CNNs) have been widely used for detecting roads~\cite{mnih2010learning}, buildings~\cite{bittner2018building}, agricultural fields~\cite{waldner2020deep}, and performing building damage assessment~\cite{robinson2023rapid} from satellite imagery. However, such approaches bring novel challenges and have required extensive resources to execute at global scales in the past --- for example, a previous effort to create a dataset of global solar photovoltaic installations from satellite imagery required processing 550 TB of imagery, used 1 million CPU-hours, and 20,000 GPU-hours~\cite{kruitwagen2021global}. Training models that generalize over the variety of landscapes on Earth requires large representative labeled datasets, running such models over terrabytes of imagery requires concentrated engineering efforts, and filtering false positives in the resulting output to produce usable data products requires non-trivial labor investments.

In this work we aim to address these challenges and present a new global map of commercial solar photovoltaic (PV) installations and onshore wind turbines based on high-resolution satellite imagery from 2024 Q2 (Figure \ref{fig:1}) as well as per feature construction dates back through October 2017. Specifically, we propose a data-centric machine learning~\cite{roscher2023data} approach to modeling solar PV and windmills from annotations sourced from OpenStreetMap along with a simple false positive filtering model based on MOSAIKS  features~\cite{rolf2021generalizable}. We test our methods on public benchmark datasets and use them to train models that we apply globally and temporally (Figure \ref{fig:2}). We validate our global map by comparing aggregate power capacity estimates at a country level to IRENA's latest 2023 per-country estimates of renewable energy (Figure \ref{fig:irena_results}). Finally, we tie each feature to the land cover that preceded it using global 300-m resolution land cover data from the European Space Agency~\cite{bontemps2015multi}. This, for the first time, provides a detailed understanding of the land use change associated with wind and solar development.

\section{Materials and Methods}

We produce global, temporal datasets of solar photovoltaic (solar PV) power plants and onshore wind turbines by training and applying semantic segmentation models on PlanetScope basemap satellite imagery every three months from October, 2017 (2017 Q4) through June, 2024 (2024 Q2). We use these models to generate two global datasets: geospatial polygons that represent individual commercial, industrial, or utility scale solar PV installations and geospatial point data that represent individual wind turbines. Specifically, we follow the convention set in Kruitwagen et al.~\cite{kruitwagen2021global} for solar -- targeting installations in excess of 10,000 square meters --- and for wind we target any onshore turbine visible at a $4.7\text{px/m}$ image resolution (which we estimate to be a windmill with a rotor diameter of 47 meters or larger). This would include most wind projects built in the last 40 years~\cite{enevoldsen2019examining}. We estimate each feature's date of construction based on when it first appeared in the satellite imagery --  already present by 2017 Q4, or quarterly from 2018 Q1 onward -- and, finally, join the features to the Copernicus Climate Change Service (C3S) global land cover dataset from 2018~\cite{copernicus2019land} to estimate the land cover type that each newly-constructed feature replaced. Fundamental to our approach is a data-centric machine learning method~\cite{roscher2023data,zha2025data} for \textit{cleaning} labels from OpenStreetMap to use for training the semantic segmentation models. Our methods can therefore be broken down into data-centric model development, global inference (i.e. dataset generation), and validation as described in the following three sections.

\subsection{Model development}

Our model development process involves sourcing globally distributed labeled data of known solar PV installations and windmills from OpenStreetMap, joining these satellite imagery appropriate to the date at which the labels are collected, cleaning these labels with an iterative data-centric ML pipeline, training semantic segmentation models to be applied globally, and finally training filtering models to remove false positives.

\subsubsection{Datasets} \label{sec:data}

\begin{figure}[!thb]
    \centering
    \includegraphics[width=1\linewidth]{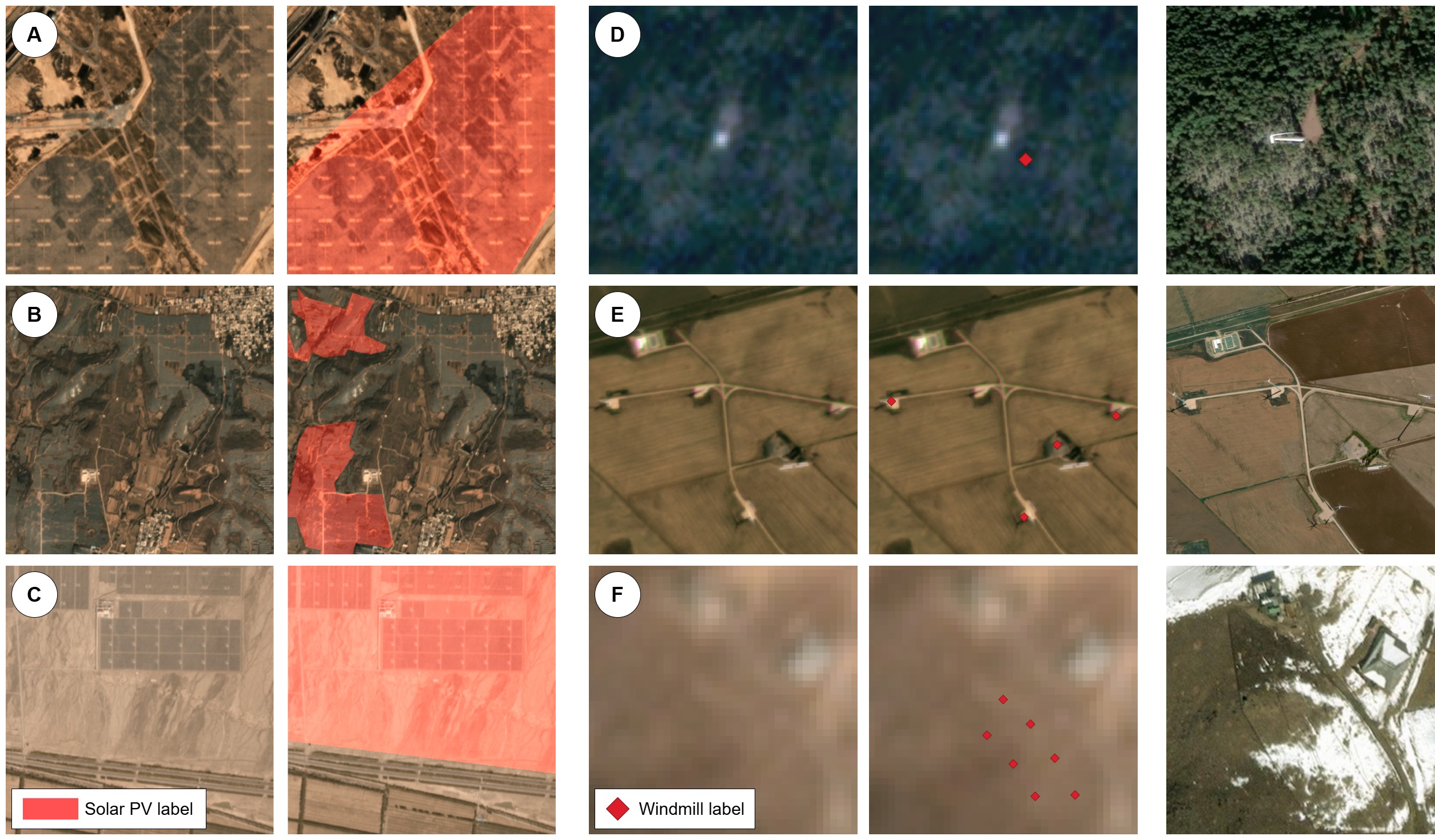}
    \caption{Examples of noisy labels of solar PV installations and windmills from OpenStreetMap. Solar PV installations may be labeled according to the \textit{total area} of the larger installation they are in (\textbf{A, C}) or might not be aligned with satellite imagery in locations where rapid expansion of renewable energy is occurring (\textbf{B}). Similarly, windmills point labels may be misaligned with satellite imagery (\textbf{D}), missing from wind farms that are in development (\textbf{E}) or out-of-date (\textbf{F}). Deep learning models trained on such labels will not generalize well as they overfit to the noise.}
    \label{fig:label_errors}
\end{figure}

\paragraph{OpenStreetMap labels}
OpenStreetMap (OSM) is a collaborative, crowd-sourced mapping platform that provides freely accessible geospatial data, widely used for scientific research, disaster response, and urban planning~\cite{haklay2008openstreetmap}. Its community-driven approach results in continuous updates and diverse spatial coverage, however also presents challenges when using the annotations as labels in machine learning pipelines. For example, Kruitwagen et al. find that the solar PV polygons that they source from OpenStreetMap have multiple meanings. Approximately 9\% of the polygons are labels of \textit{total area} (i.e. ``the full area of the installation facility''), 18\% are labels of \textit{direct area} (i.e. ``the area covered by the solar arrays, land in between them, and supporting equipment''), and 73\% annotate the \textit{array area} (i.e. only the area containing solar panels). Further, any spatial or temporal misalignments between the high-resolution satellite imagery basemaps that the OSM contributors use to annotate the features and the satellite imagery used in modeling will introduce noise (i.e. incorrect labels). Figure \ref{fig:label_errors} shows six examples of such semantic, spatial, and temporal noise.

Regardless, OSM is a valuable source of information to begin the modeling process. Kruitwagen et al. model global solar PV installations in 2018 based on a training set of OSM derived solar PV labels and a mixture of Sentinel-2 and SPOT satellite imagery~\cite{kruitwagen2021global}. Further, they release training, validation and testing polygon data used in their modeling approach~\cite{kruitwagen2021Dataset}. Dunnet et al. created and published a harmonized dataset of solar PV and windmill installations based exclusively on OSM data from 2020~\cite{dunnett2020harmonised}. Specifically, they query OpenStreetMap with a variety of different key/value tag combinations, find a total of 326,234 solar polygons, 1,808,585 solar points, 1,889 wind polygons, and 305,306 wind points, then cluster these into installation level groupings. Finally, they estimate installation level power capacity using a regression model and argue that the final dataset is spatially representative.

We use the solar training, validation, and test datasets from Kruitwagen et al. and the windmill point dataset from Dunnet et al. to develop and test dataset cleaning methods. The Kruitwagen training dataset consists of 18,502 samples with OSM based labels. The validation and testing datasets consist of \textit{direct area} labels generated from manual annotation of Sentinel-2 and high-resolution basemaps from 2018 Q4. We pair all labels from Kruitwagen et al. with PlanetScope imagery from 2018 Q4 using the areas from the respective train, validation and test tile definitions found in the data release~\cite{kruitwagen2021global}. For wind, we use the 272,503 windmill point dataset in the Dunnet et al. final data release, then perform a 2-stage DBSCAN clustering to first cluster the points into local groups (using a 200m radius) then second cluster the groups of points into 17,971 non-overlapping \textit{tiles} (using a 2000m radius). We partition the 17,971 tiles into training, validation, and test sets randomly with 80/10/10 proportions respectively. Finally, we pair each tile with PlanetScope basemap imagery from 2020 Q4, i.e. the latest time period that the labels were collected.

\paragraph{PlanetScope basemaps}
PlanetScope RGB visual basemaps are color-corrected mosaics generated from daily imagery captured by the PlanetScope satellite constellation sampled at a spatial resolution of 4.7m/px at the equator\footnote{This corresponds to the spatial resolution zoom level 15 in the Web Mercator (EPSG:3857) slippy map tile specification at the equator.}~\cite{planetscope2024}. To construct the basemaps, Planet uses a proprietary ``best scene on top'' mosaicking algorithm that prioritize the highest quality individual PlanetScope scenes over a given time period based on cloud cover, viewing angle, and radiometric stability. These selected scenes are blended into a composite and normalized according to MODIS monthly surface reflectance values to provide a seamless image, then distributed as a uniform grid of $4096 \times 4096$ pixel \textit{quads}~\cite{planetspecs2023}. We use the quarterly basemap product from 2017 Q4 to 2024 Q2 in both our solar and wind modeling pipelines.

\begin{table}[thb]
\centering
\resizebox{\textwidth}{!}{%
\begin{tabular}{@{}cccccccc@{}}
\toprule
 & \multirow{2}{*}{\textbf{Method}} & \multicolumn{3}{c}{\textbf{Pixel level}} & \multicolumn{3}{c}{\textbf{Object level}} \\ \cmidrule(lr){3-5} \cmidrule(lr){6-8}
 &  & \textbf{Precision} & \textbf{Recall} & \textbf{F2 Score} & \textbf{Precision} & \textbf{Recall} & \textbf{F2 Score} \\ \midrule
\multirow{5}{*}{\textbf{Solar}} & No filtering & 21.21\% & 83.49\% & 52.60\% & 6.53\% & 22.90\% & 15.26\% \\
 & 1 iteration & 45.17\% & 72.44\% & 64.63\% & 22.61\% & 20.65\% & 21.02\% \\
 & 2 iterations & 45.98\% & 80.35\% & 69.90\% & 30.33\% & 21.15\% & 22.51\% \\
 & 2 iterations + hard negatives & 57.19\% & 80.59\% & 74.50\% & 23.97\% & 21.62\% & 22.05\% \\
 & 2 iterations + hard negatives + FP model & 57.73\% & 80.31\% & 74.48\% & 50.93\% & 19.60\% & 22.35\% \\ \midrule 
\multirow{2}{*}{\textbf{Wind}} & No filtering & - & - & - & 59.63\% & 71.48\% & 68.75\% \\
 & Global model + filtering + hard negatives & - & - & - & 90.81\% & 81.63\% & 83.31\% \\ \bottomrule
\end{tabular}%
}
\caption{Pixel and object level performance results for the  solar PV and windmill detection models on the test sets of the Kruitwagen et al.~\cite{kruitwagen2021global} and Dunnet et al.~\cite{dunnett2020harmonised} datasets with and without data and prediction cleaning methods.}
\label{tab:results}
\end{table}

\subsubsection{A data-centric approach to modeling wind and solar} \label{sec:filtering}

We model solar PV installations as a two class semantic segmentation problem (``background'' and ``solar PV'') followed by a vectorization step using a U-Net~\cite{ronneberger2015u} based fully convolutional model with a ResNext-50 backbone~\cite{xie2017aggregated} pre-trained on ImageNet. We optimize using a weighted cross entropy loss (weighting the ``background'' and ``solar'' classes with 0.3 and 0.7 respectively, determined through offline experimentation) and the AdamW optimizer~\cite{loshchilov2017decoupled} with a learning rate of $1e^{-4}$. We use an image size of $256 \times 256$ pixels and apply random rotation, flipping, color jitter, and sharpness augmentations. Similarly, we model wind turbines as a two-class (``background'' and ``wind turbine'') point-based semantic segmentation task using a Fully Convolutional Network (FCN)~\cite{long2015fully} with a ResNet-50 backbone. We optimize the wind turbine model using the localization-based counting (LC) loss proposed by Laradji et al.~\cite{laradji2018blobs} with four terms: a ``patch-level loss'', ``point-level loss'', ``split-level loss'', and ``false positive loss'' (See Section \ref{sec:wind_loss} in the Appendix for more details). We train the wind turbine model with the AdamW optimizer, a learning rate of $1e^{-5}$, a patch size of $128 \times 128$ pixels, and rotation and random crop augmentations.

Given a training dataset, $\{(X_i, Y_i)\}_{i=1}^N$, of $N$ satellite imagery patches, $X$, and associated solar PV or windmill labels as pixel masks, $Y$, our label cleaning process involves iteratively training models $f(X_i) = \hat{Y}_i$ using the modeling methods described above and filtering out training samples that the model is unable to fit. While deep learning models are overparameterized and often have the capacity to ``memorize'' training samples, convolutional neural networks in particular have strong texture biases~\cite{geirhos2018imagenet} and will have difficulty fitting contradictory pixel level labels over similar inputs. For example, if desert/barren land in satellite imagery is typically labeled as ``background'' throughout the training set, but labeled as ``solar PV'' in a few samples due to the label noise described in the Figure \ref{fig:label_errors}, then the model will have to learn spurious long range features to (in)correctly fit those noisy training samples. In doing so, the model is likely to make mistakes in similar settings -- for example, predicting the ``solar'' class in a large buffer around every installation. The intuition behind our method follows this observation -- if a model is unable to overfit to a label (under normal amounts of regularization), then that label is likely incorrect or, at the least, contradicting other labels, and should be removed. We apply this by measuring $\text{IoU}(Y_i, \hat{Y}_i)$ per sample and removing the sample from the dataset if it is lower than a threshold value of $0.1$. We repeat this process, retraining the model after each pass over the training dataset, until all samples can be fit with $\text{IoU} > 0.1$. We additionally remove all training samples where over 90\% of the mask is labeled as solar PV as this leads to high bias values in the final layers of the model.

\paragraph*{Hard negative mining}
It has been shown that some form of hard negative mining is useful to improve the performance of object detectors~\cite{jin2018unsupervised,zhao2019object,ortiz2022artificial}. This is especially relevant for object detectors applied over large amounts of satellite imagery as the number of false positives scales with the amount of imagery that a model is run on. As such, we adopt a bootstrapping~\cite{wan2016bootstrapping} approach where we train an initial model, run inference with it across satellite imagery from different geographies, then visually inspect the predictions for false positives. We add false positive predictions back to the training set of the wind or solar models as ``hard negative samples'' for further training runs.

\paragraph*{Filtering false positives}
Finally, to further improve precision, we implement a two-stage modeling approach similar to Kruitwagen et al.~\cite{kruitwagen2021global} and Robinson et al.~\cite{robinson2022mapping} where we train an additional model to filter false positive predictions from the set of predictions made by our initial model. The intuition behind this approach is that the initial model can maximize recall -- finding all relevant instances of the object of interest at scale from imagery layers -- while the filter model can remove false positives, potentially using accessory information (e.g. shape level features) that would be difficult or impractical to include in the first model. Specifically, we train a SVM that uses 256 MOSAIKS~\cite{rolf2021generalizable} random convolutional features pooled over the pixels in an object and labels collected in an online fashion. We reweight the samples and tune the margin-loss of the SVM to be precise (i.e. to remove false positives without removing true positives) instead of sensitive.

We experiment with these methods in the context of the Kruitwagen et al. and Dunnett et al. benchmark datasets in Section \ref{sec:results}.

\subsubsection{Global temporal inference}

We run our final solar PV and wind segmentation models on the entire 2024 Q2 PlanetScope basemap. This data is split into 833,712 $4096 \times 4096$ pixel \textit{quads} -- approximately 13.98 trillion pixels -- and occupies 38.42 TB on-disk. We run our models in parallel on virtual machines in the same cloud datacenters that the imagery is stored in and achieve a throughput of 1.16 quads / second for the solar model and 0.51 quads / second for the wind model on 16GB V100 GPUs resulting in a total cost of 658.1 V100 GPU hours\footnote{This translates to a cost between \$281 and \$2014 USD of compute using Azure ``spot'' and ``pay-as-you-go'' pricing respectively as of March 2025.}. We vectorize the resulting dataset and perform false positive filtering as described above, resulting in a final dataset of 375,197 wind turbines and 86,410 solar PV installations. The wind detection model predicts small blobs over individual wind turbines, hence centroids are extracted from the polygonization of the blobs for the final wind turbines dataset. 

\paragraph{Construction dates}
For each of the above features, we run the models on the quarterly time series of PlanetScope basemap imagery cropped to the feature's extent starting in 2017 Q4. This is a negligible cost compared to the initial global inference and record the first date that the model has positive predictions consisting of at least 10\% of the area as the construction date.

\paragraph{Land cover features} We record the land cover of each feature in 2018 using the ESA CCI's medium-resolution land cover (MRLC) dataset. For solar we record the mode land cover intersecting with the polygon prediction and for wind we record the land cover of the pixel intersecting with the point label. This dataset gives annual land cover maps with 22 land cover classes at a 300 m/px spatial resolution from 1992 to 2020~\cite{harper202329,defourny2023observed}.

\begin{figure}[t]
    \centering
    \includegraphics[width=0.8\linewidth]{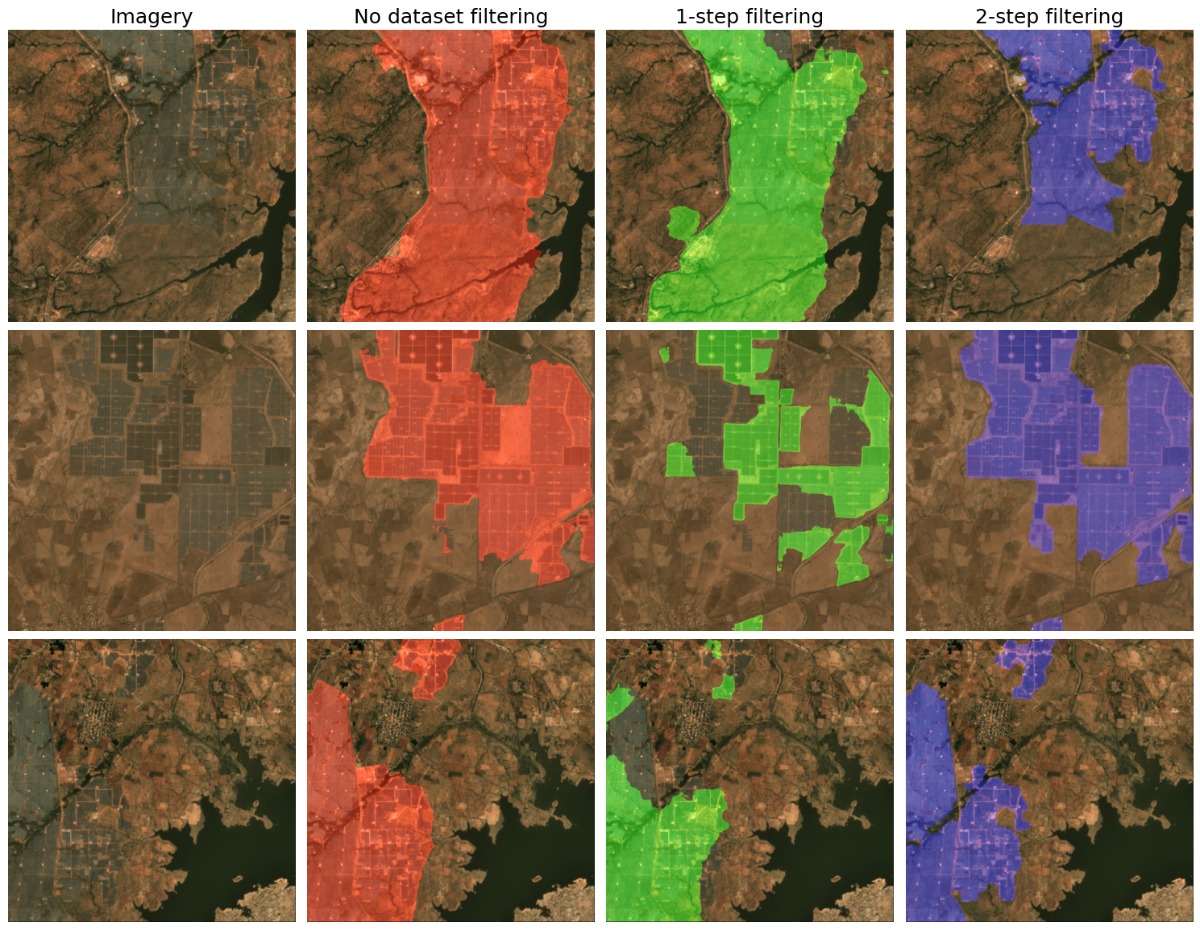}
    \caption{Progression of the predictions from a solar PV segmentation model trained on unfiltered versus filtered datasets.}
    \label{fig:solar-progression}
\end{figure}

\section{Experiments and Results}
\label{sec:results}

\subsection{Data-centric experiments} \label{set:comparetoGT}

We evaluate the impact of our proposed data cleaning, hard negative mining, and false positive filtering method steps on the Kruitwagen et al. and Dunnett et al. benchmark datasets described in Section \ref{sec:data}.

For solar, the Kruitwagen training set originally contains 18,502 samples. Applying the filtering methodology (described in Section \ref{sec:filtering}) results in an initial reduction to 17,832 samples, then again to 17,152 samples. Table \ref{tab:results} shows this results in an increase in pixel level F2 from 52.6\% to 69.9\%. The addition of 10,289 hard negative samples further improves pixel-level performance, raising the F2 score to 74.5\%. Finally, the addition of the false-positive filter model --- trained with 300 (~5\%) labeled predicted polygons --- maintains the overall pixel-level F2 score (74.48\%) while significantly improving object-level precision, increasing it from 23.97\% to 50.93\% (while lower object level recall from 21.62\% to 19.60\%). Qualitatively, the predictions from models trained on the filtered dataset produce segmentation masks that better delineate the \textit{direct area} of solar panels in an installation. Figure \ref{fig:solar-progression} shows the progression of the predictions of models trained on the three datasets across three large installations. We observe that simply pairing OSM solar generator labels with satellite imagery is not sufficient for training generalizable models, but instead results in oversegmentation of solar PV installations, large numbers of false positives, and other inference artifacts when applied at scale (\textbf{column 2}). Progressively filtering out training samples that a U-Net is unable to overfit to results in models with better performance (\textbf{columns 3 and 4}) under the same model architecture and training methods. 

\begin{figure}
    \centering
    \includegraphics[width=1\linewidth]{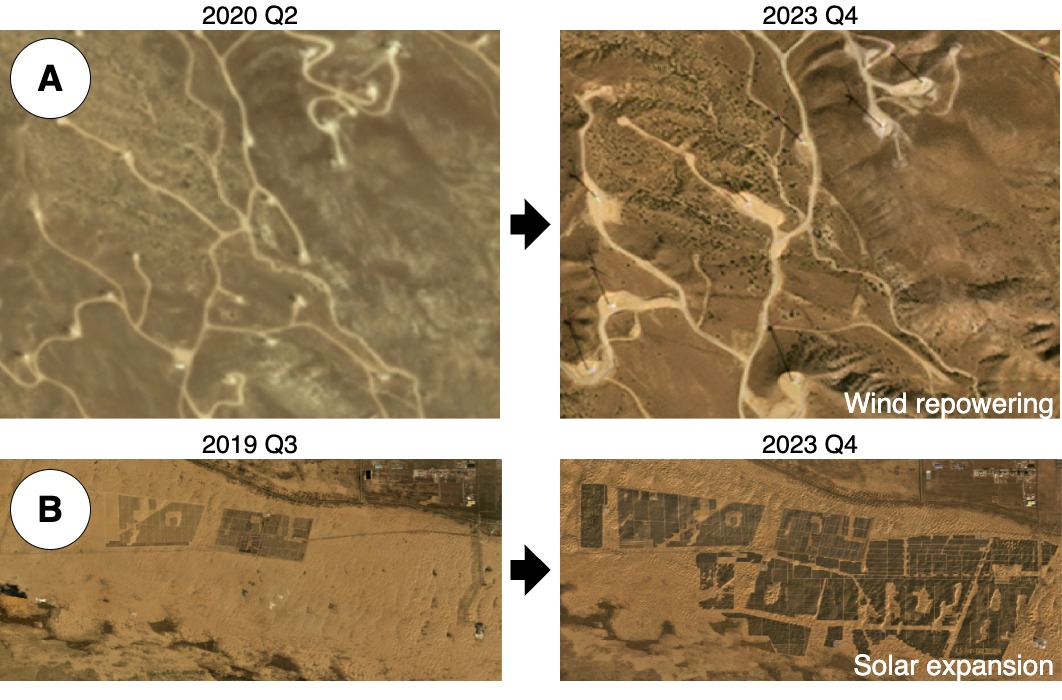}
    \caption{Effect of repowering a wind farm in Clear Lake, Iowa, United States (\textbf{A}) and solar expansion from 2019 Q3 to 2023 Q4 in Ordos City, China (\textbf{B}). Repowering of wind farms involves replacing older, less efficient wind turbines with newer, more powerful models to increase energy production and extend the lifespan of a wind farm, often using existing infrastructure. Solar farms footprints also tend to grow rapidly over time. These are common practices that make labels for single time stamp obsolete over time. Repowering along with continuous and rapid development of new solar and wind installations are some of the factors that make temporal monitoring crucial.}
    \label{fig:wind_solar_changes}
\end{figure}

For windmill detection, applying a global model with hard negative mining results in substantial improvements at the object level, with precision increasing from 59.63\% to 90.81\% and F2 score improving from 68.75\% to 83.31\%. We note that these numbers are likely underestimating performance due to quality issues with OSM labels which are often misaligned, mislabel of individual turbines, or the repowering of wind farms as shown in Figure \ref{fig:label_errors} and Figure \ref{fig:wind_solar_changes}.

\subsection{Country-level capacity estimates evaluation} 

\begin{table}[!t]
\centering
\begin{tabular}{@{}ccccc@{}}
\toprule
\textbf{Dataset} & \textbf{Comparison Layer} & \textbf{Kendall's $\tau$} & \textbf{Pearson's $r^2$} & \textbf{Global capacity difference (GW)} \\ \midrule
Ours             & Solar PV     & 0.718 & 0.960 & -574.5 \\
Satlas           & Solar PV     & 0.676 & 0.945 & -767.7 \\
OSM              & Solar PV     & 0.683 & 0.478 & -805.4 \\ \midrule
Ours             & Onshore wind & 0.877 & 0.932 & 181.4  \\
Satlas - Onshore & Onshore wind & 0.804 & 0.978 & 51.2   \\
OSM - Onshore    & Onshore wind & 0.819 & 0.898 & 188.4  \\ \midrule
Satlas           & All wind     & 0.792 & 0.975 & 12.9   \\
OSM              & All wind     & 0.793 & 0.893 & 153.9  \\ \bottomrule
\end{tabular}%
\caption{Comparison of country-level capacity estimates derived from different global solar PV and wind layers to IRENA's 2023 country level capacity estimates. The global capacity difference is negative where the individual layers underestimate capacity and positive where they overestimate capacity.}
\label{tab:irena}
\end{table}

\begin{figure}[t]
    \centering
    \includegraphics[width=0.45\linewidth]{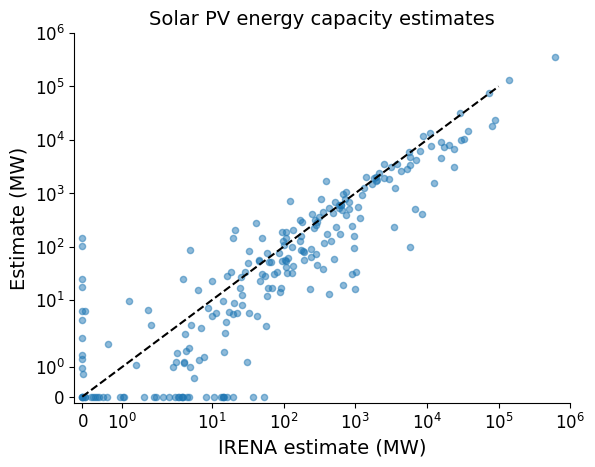}
    \includegraphics[width=0.45\linewidth]{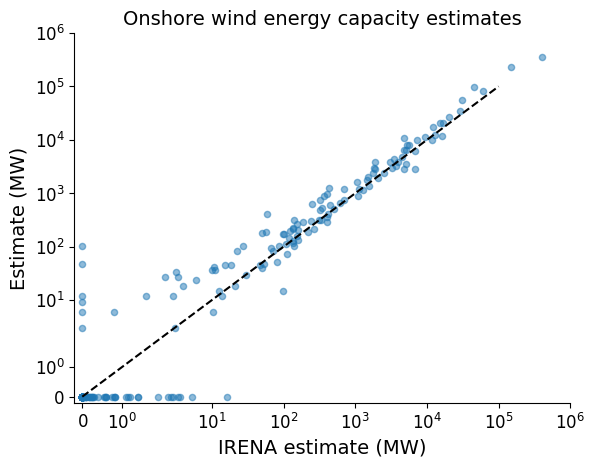}
    \caption{Aggregate country level capacity estimates from our estimates compared to IRENA country level capacity estimates for 2023.}
    \label{fig:irena_results}
\end{figure}

The International Renewable Energy Agency (IRENA) publishes an annual Renewable Capacity Statistics report, which provides global and country-level data on installed renewable energy capacity. Estimates include hydro, onshore and offshore wind, solar photovoltaic (PV) and concentrated solar power (CSP), bioenergy, geothermal, and marine energy. The agency compiles data from various sources, including government reports, utility data, industry surveys, and grid operators to compile their capacity estimations. In cases where official data is not available, IRENA uses statistical modeling and extrapolation based on historical trends, industry reports, regional energy demand, company disclosures, and project announcements~\cite{2024irenam}.

We validate our solar PV and onshore wind farm predictions by aggregating them to the country level, estimating total capacity based on the construction date estimates, total square kilometers of solar PV and number of wind turbines, then comparing those estimates to the most recent Renewable Capacity Statistics report from IRENA corresponding to 2023. 

Specifically, for solar we linearly interpolate a 2013 solar power density estimate of 30 MW/sq km (Ong et al~\cite{ong2013land}) and a 2022 estimate of 69 MW/sq km (Bolinger and Bolinger~\cite{bolinger2022land} for tracking plants) to derive coarse power density values that depend on construction date. Specifically, for solar installations built before 2019 we assume 51.5 MW/sq km, then assume power density increases by $\approx$ 4.33 MW/sq km per year through 2022 (using a final value of 65.9 for 2022, 2023, and 2024). We note that this is a \textit{coarse} estimate of capacity which depend per-installation on additional factors such as fixed-tilt vs. tracking, inverter load ratios, and panel construction materials. For wind we assume a fixed 3 MW per turbine following the average wind turbine capacity over time estimated by the US Department of Energy based on the average turbine size and rotor diameter~\cite{2024doe}.

We apply this country-level capacity estimation method to our global dataset, wind and solar generator plants from OSM\footnote{Following the methodology of Dunnett et al.~\cite{dunnett2020harmonised}.} in January 2024, and the Satlas renewable energy layers~\cite{bastani2023satlaspretrain}. Table \ref{tab:irena} show Kendall's $\tau$ rank order correlation, Pearson's $r^2$, and the difference in global capacity from IRENA's 2023 estimates. Our dataset shows the strongest agreement with IRENA for solar PV, achieving the highest Kendall's $\tau$ (0.718) and Pearson's $r^2$ (0.960), with a global underestimation fo 574.5 GW. The Satlas and OSM datasets show lower correlations with OSM in particular having the largest underestimation and lowest correlation (0.478). For onshore wind, our dataset also demonstrates a high Kendall's $\tau$ (0.877) indicating the windmill predictions are distributed across countries proportional to their estimated onshore wind energy capacity. Satlas achieves the highest Pearson's $r^2$ (0.978) and closest match to the total capacity. OSM's wind data performs comparably in terms of Kendall's $\tau$ (0.819) but has a slightly weaker Pearson's $r^2$ (0.898). Across all wind energy sources, Satlas yields the best Pearson's $r^2$ (0.975) and the smallest deviation from IRENA estimates (+12.9 GW), suggesting it captures overall wind capacity trends effectively. Figure \ref{fig:irena_results} shows scatter plots for both our wind and solar capacity country level estimates.

\section{Usage Notes}

The Global Renewable Watch provide some of the first globally consistent land development patterns at a high resolution ($4.7\text{m/px}$ ) that depict the patterns of wind and solar development. 
Our approach offers an advancement to other products by including both wind and solar development patterns, as well as factoring in resource yield potential that can be used to estimate installed capacity. It also advances the global mapping of the growing energy footprint that increasingly play a role in land use change \cite{oakleaf2019mapping}, but have been overlooked relative to agriculture or urban expansion. Further, we examined the performance and sensitivity of both wind and solar model outputs and validated results with the best available known locations of recent or planned development. 

We acknowledge that our AI based models, like all other similar products, are inherently prone to inaccuracies, omissions, and inconsistencies in both their spatial features and attributes. While we used the best publicly available and current data for model training, input datasets were not always comprehensive in regional coverage (an issue that plagues all global analyses). The latter may influence accuracy of the AI models at local scales. 

Despite these limitations, the timeliness and substantial need for these types of data is clearly demonstrated by an increasing number of online portals hosting data on renewable energy development patterns: e.g., the World Resource Institute's (WRI) \href{https://www.wri.org/research/state-nationally-determined-contributions-2022}{The State of Nationally Determined Contributions}, \href{https://www.climatewatchdata.org/net-zero-tracker}{ClimateWatch's Net-Zero Tracker}, \href{https://ember-climate.org/insights/research/global-electricity-review-2023/#supporting-material}{Ember's Global Electricity Review}, among others. We note that the renewable development patterns on these online portals predominately focus on the amount of renewable energy, i.e., installed or production capacity. Thus, they are limited to assessments that seek to understand the mix and amount of renewable energy and as a result have limited application to exercises that seek to guide planning of future energy development. Of the select datasets that assess renewable energy patterns, they tend to map only amount of renewable energy summarized at the national, state or other coarse scale jurisdiction boundaries without integrating spatial feasibility factors. These products capture resource yield but without spatial details on siting constraints or siting feasibility that can be used to guide future planning exercises.

We see three primary applications enabled by providing complete and up-to-date locations of wind and solar installations. First to serve as a complimentary data source that can be used to estimate installed capacity of wind and solar development. Given that we are using area estimates for solar and individual wind turbine capacity averages for wind, these estimates are likely to provide more supportive information to the more accurate accounting provided by other global (e.g. IRENA) and country-level (e.g. US EIA) entities. To ensure reporting is accurate however these data can be used to provide average capacity values per area or turbine to check if these estimates fall within current acceptable power densities. Second our dataset can help identify where development is occurring at the expense of conservation assets and smart siting approaches are most needed. This early warning can help guide where limited planning resources can be used most effectively. Additionally with the temporal nature of these data, we can now track progress of implementing effective and proactive planning. As wind and solar investments scale up it will be critical to develop plans that can guide the identification of locations to meet these growing investments. Finally, by using the spatial patterns of past development, we can create predictive models that not only identify future conflicts but also help identify where these RE demands can be met on low conflict land and thus further expedite development~\cite{ortiz2022artificial}.

\bibliographystyle{plainnat}
\bibliography{citations} %

\clearpage
\newpage

\appendix

\section{Windmill point-based segmentation loss terms}
\label{sec:wind_loss}

We tackle the problem of detecting individual wind turbines from Planet Labs' PlanetScope global quarterly basemap imagery as a point-based semantic segmentation task. Our windmill detection model consists of a Fully Convolutional Network (FCN) ~\cite{long2015fully} with a ResNet-50 backbone trained using the localization-based counting (LC) loss original proposed by Laradji et al.~\cite{laradji2018blobs} for the task of object counting and we adapt it to segment and detect windmills.  We train our wind turbine detector with a four component LC loss: 
\begin{enumerate}
    \item \textit{Patch-level loss}, is a negative log-likelihood loss which encourages the model to detect wind turbines at the image patch level. This is a global loss component that encourages the model to predict at least a pixel as with turbines for image patches with windmills labels on it, while penalizing the model for predicting any windmill pixels for patches with no turbines.
    \item \textit{Point-level loss}, serves as a localization loss. It encourages the model to predict wind turbines for the pixels matching the point based windmill annotations by computing a negative log-likelihood loss only to the annotated pole pixels. Both the patch and point level losses evaluate the likelihood that a pixel is in the pole class. The next two loss components, evaluate the likelihood that a pixel belongs to the background (anything that is not a wind turbine) class.
    \item \textit{Split-level loss} is a loss component which encourages the model to predict unique blobs for each wind turbine. First, boundaries between ground truth poles are found using a watershed segmentation algorithm~\cite{beucher2018morphological}. These boundaries are annotated as background while the area within the boundaries are annotated as foreground in an online fashion while training. The model learns to predict the probability that a pixel belongs to the background class. This is also done by computing a negative log-likelihood loss weighted by the number of windmills within the blob. This encourages the model to predict each individual turbine as a blob.
    \item \textit{False positive loss} The last component discourages the model from making false positive predictions. A negative log-likelihood loss is applied to pixels with no turbine annotations to minimize the false positive predictions by the model. This loss component was up-weighted for hand-labeled energy transmission and telecommunication towers hand-labeled in the training set and found to be false positives. 
\end{enumerate}

\end{document}